\documentclass{article}
\usepackage{arxiv}

\usepackage{subfigure}
\usepackage[utf8]{inputenc} 
\usepackage[T1]{fontenc}    
\PassOptionsToPackage{hyphens}{url}
\usepackage{hyperref}       
\usepackage{microtype}      

\usepackage{amsmath,graphicx,amssymb,amsfonts,nicefrac,amstext}
\usepackage{pgfplots}
\pgfplotsset{compat = 1.3}
\usepackage{multirow}
\usepackage{makecell}
\usepackage{booktabs}
\usepackage{tabularx}
\usetikzlibrary{pgfplots.groupplots}
\usepackage{numprint}
\usepackage{color, colortbl}
\usepackage{natbib}
\title{Rethinking Evaluation in ASR: \\Are Our Models Robust Enough?}

\author{Tatiana Likhomanenko\thanks{Equal contribution.}\\
Facebook, Menlo Park\\
\begin{small}\texttt{antares@fb.com}\end{small} \And Qiantong Xu$^*$\\
Facebook, Menlo Park \\
\begin{small}\texttt{qiantong@fb.com}\end{small} \And Vineel Pratap$^*$\\
Facebook, Menlo Park \\
\begin{small}\texttt{vineelkpratap@fb.com}\end{small} \And Paden Tomasello\\
Facebook, Menlo Park \\
\begin{small}\texttt{padentomasello@fb.com}\end{small} \And Jacob Kahn\\
Facebook, Menlo Park \\
\begin{small}\texttt{jacobkahn@fb.com}\end{small} \And Gilad Avidov\\
Facebook, Menlo Park \\
\begin{small}\texttt{avidov@fb.com}\end{small} \And Ronan Collobert\\
Facebook, Menlo Park \\
\begin{small}\texttt{locronan@fb.com}\end{small} \And Gabriel Synnaeve\\
Facebook, Paris \\
\begin{small}\texttt{gab@fb.com}\end{small}}
\hypersetup{
pdftitle={Rethinking Evaluation in ASR: Are Our Models Robust Enough?},
pdfauthor={Tatiana Likhomanenko, Qiantong Xu, Vineel Pratap, Paden Tomasello, Jacob Kahn, Gilad Avidov, Ronan Collobert, Gabriel Synnaeve},
pdfkeywords={automatic speech recognition, generalization, transfer, robustness},
}

\begin{document}

%
\maketitle
\begin{abstract}
Is pushing numbers on a single benchmark valuable in automatic speech recognition? Research results in acoustic modeling are typically evaluated based on performance on a single dataset. While the research community has coalesced around various benchmarks, we set out to understand generalization performance in acoustic modeling across datasets~--~in particular, if models trained on a single dataset transfer to other (possibly out-of-domain) datasets. 
We show that, in general, reverberative and additive noise augmentation improves generalization performance across domains.
Further, we demonstrate that when a large enough set of benchmarks is used, average word error rate (WER) performance over them provides a good proxy for performance on real-world noisy data. Finally, we show that training a single acoustic model on the most widely-used datasets -- combined -- reaches competitive performance on both research and real-world benchmarks.
\end{abstract}

\section{Introduction}
\label{sec:intro}
Progress in automatic speech recognition (ASR) is measured on the validation and test sets of standard datasets. However, most acoustic models (AMs) are often developed and tuned on a single dataset and transfer poorly to other datasets. Moreover, most large standard benchmarks have similar domains and recording conditions, often with little background noise or little reverberation.
These factors lead to siloed ASR research.
Benchmarks in noisy conditions exist (e.g.~\citep{barker2018fifth,maciejewski2020whamr}), but are limited in training set size.
A unified benchmark comprised of conversational, oratory, and read speech with varied recording conditions and noise would certainly serve the research community well; here, however, we study how the currently-popular public benchmarks can be used to gauge model generalization performance.

Our approach constructs a validation procedure -- using only public datasets -- that is a better predictor of overall and domain transfer performance than datasets taken in isolation. We train the same state-of-the-art model architecture on different benchmarks pushing for best performance on each benchmark separately. We also jointly train a model on all datasets. Given the transfer performance on test sets, we can ascertain which test sets are good proxies for transfer performance as well as which training sets can produce the best-performing models.
Additionally, we train models with additive noise of signal noise ratios (SNR) and evaluate performance on the aforementioned validation sets.
This informs us on the robustness of various datasets in transfer and which test sets 
are the best predictors of ASR performance in others. Finally, we look at the performance, in transfer only, on our in-house ASR datasets. This informs us about which sets of test sets should be used if one wants to transfer to a wide range of conditions of speech.

\section{Related Work}
\label{sec:relatedwork}
Previous works that study transfer in ASR include \citep{ghahremani2017investigation} that studied transferring varying number of layers trained out-of-domain, from SwitchBoard to AMI-IHM or from LibriSpeech to AMI-IHM. In this paper as in ours, a joint model trained on multiple out-of-domain datasets exhibits better transfer. In the context of the Arabic MGB-3 challenge, \citet{manohar2017jhu} transfered AMs trained on broadcast TV to Youtube videos, with a different setting than here as the training transcriptions were noisily labeled. Distillation was used to improved transfer in~\citep{asami2017domain}, where the soft-target part of the distillation loss may help with regularization. For another kind of transfer in~\citep{kunze2017transfer}, the authors transferred LibriSpeech trained wav2letter~\citep{collobert2016wav2letter} models to German by fine-tuning them on German, with better performance than training from scratch. Very recently, \citet{szymanski2020we} point out some limitations of current ASR benchmarks, and propose guidelines to create multi-domain datasets. Finally, while DeepSpeech~2~\citep{deepspeech2} did not focus their study on transfer, we train a single AM on multiple datasets at once, as they did.

We also explore how training with additive noise helps transfer (on clean and on noisy conditions). Some of the first studies of noise robust ASR with deep networks include \citep{vinyals2012revisiting,seltzer2013investigation} which respectively trained RNNs on Aurora-2 and DNNs on Aurora-4 (a classic noise-robust ASR benchmark). The first looks at the performance in transfer of an RNN-based acoustic model trained on clean speech only, while the second investigates different regimes of noise-aware training for DNNs and which are most beneficial. A top entry (with models ensembling) from Chime-4 \citep{menne2016rwth} compares different beamforming approaches for far-field ASR. More recently, Chime-5 \citep{barker2018fifth} established a benchmark for far field ASR with background noise that is labeled, and WHAMR! \citep{maciejewski2020whamr} is another benchmark with additive noise and reverberant speech synthetically mixed out of good quality noise samples and simulated room impulse responses (RIRs). On this topic, a paper \citep{ko2017study} showed that -- when done properly -- we can use simulated RIRs, by comparing their influence to real room impulse responses in far-field ASR performance. Finally, we do not study more advanced data augmentations than additive noise and reverberation, but their impact on transfer is an avenue for future research. For instance, the authors of \citep{sun2018training} perform adversarial data augmentation through a Fast Gradient Sign Method attack on the current model's parameters, which leads to consistent gains on Chime-4 and Aurora-4.

\section{Domain Transfer}
\label{sec:domaintransfer}

In order to study transfer across datasets and conditions, we do a systematic analysis. 
In all our experiments, we use a single Transformer-based AM architecture with 270M parameters, to make our results comparable across the board. We train multiple single-dataset baselines as well as one joint model trained on all datasets at once. We then evaluate this set of models on all the validation and test sets, to measure how each ``in-domain'' model transfers to ``out-of-domain'' datasets.
From this, we analyze which datasets suffer more acutely from ``domain overfitting.'' Evidently, it is difficult to separate the ``in-domainness'' and size of a dataset; e.g., we cannot directly compare results on WSJ (80h) to ones on LibriSpeech (960h). 
We also fine-tune our joint model with additive noise and artificial reverberation and measure how it boosts transfer (joint+noise model).
We also fine-tune our joint and joint+noise models on the transfer dataset with 10min, 1h, 10h, and 100h of in-domain data. 
Finally, we examine how our models transfer to real data and in the process observe that public validation and test sets performance is predictive of the transfer performance of a model to real data. 

\section{Experiments}
\label{sec:experiments}
\subsection{Datasets}

To measure domain transfer, we restrict experiments to use only datasets in English, for which there exist several commonly-used and publicly available datasets with hundreds hours of transcribed audio. Validation sets from each dataset are used to optimize model configurations and to perform all hyper-parameter tuning, while test sets are used for final evaluation only. 
\\
{\bf LibriSpeech (LS)}~\citep{panayotov2015librispeech} consists of read speech from audiobook recordings. We use standard split of train, validation (\emph{dev-clean}, \emph{dev-other}) and test sets (\emph{test-clean}, \emph{test-other}). 
\\
{\bf SwitchBoard \& Fisher (SB+FSH)} consists of conversational telephone speech. To create a training set, we combine Switchboard~\citep{godfrey1993switchboard} and Fisher~\citep{cieri2004fisher,cieri2004fisher2}. We use RT-03S~\citep{rt03s} as the validation set; test sets are the Hub5 Eval2000~\citep{hub5} data with two subsets, SwitchBoard (SB) and CallHome (CH). For the data processing and evaluation, we follow the recipe provided by Kaldi~\citep{daniel2011kaldi}.
\\ 
{\bf Wall Street Journal (WSJ)}~\citep{garofolo1993csr,linguistic1994csr,woodland1994large}. We consider the standard subsets \emph{si284}, \emph{nov93dev} and \emph{nov92} for training, validation and
test, respectively. We remove any punctuation tokens from \emph{si284} transcriptions when used for training.
\\ 
{\bf Mozilla Common Voice (CV)} project~\citep{ardila2020common}. The CV dataset consists of transcribed audio in various languages where speakers record text from Wikipedia. Anyone can submit recorded contributions; as a result, the dataset has a large variation in quality and speakers. We use the English dataset\footnote{June 22nd 2020's snapshot: \url{https://tinyurl.com/cvjune2020}. Transcriptions contain upper-case and non-English characters and punctuation. To have similar transcription normalization as in other datasets, we normalize the text for all splits: lower-casing, Unicode normalization, removing punctuation and non-English tokens, and mapping common abbreviations (e.g. ``mr.'' to ``mister'').}, where data splits are provided therein. 
\\
{\bf TED-LIUM v3 (TL)}~\citep{hernandez2018ted} is based on TED conference videos. We use the last edition of the training set from this dataset (v3), for which the validation and test sets are kept consistent (and thus numbers are comparable) with the earlier releases. We follow the Kaldi recipe~\citep{daniel2011kaldi} for data preparation.
\\
{\bf Robust Video (RV)} is our in-house English video dataset, which are sampled from public social media videos and aggregated and deidentified before transcription. These videos contain a diverse range of speakers, accents, topics, and acoustic conditions making ASR difficult. The test sets are composed of \emph{clean}, \emph{noisy} and \emph{extreme} with \emph{extreme} being the most acoustically challenging subset among them. 
The validation set comprises of data from \emph{noisy} and \emph{extreme} subsets.
\\
{\bf CHiME-6}~\citep{watanabe2020chime} is a noisy low resource dataset set, which contains around 40 hours of distant microphone conversational speech recognition in everyday home environments. We use this dataset only to evaluate robustness to noisy conditions of our models.
Front end enhancement for development and test sets is done following the official recipe~\citep{watanabe2020chime}: the guided source separation~\citep{boeddeker2018front} with 12 channels is used to enhance front end.  

\begin{table}[t!]
\caption{Statistics on datasets: sampling frequency, duration (in hours), and speech type. \label{tab:data}}
\begin{center}
\setlength\tabcolsep{6pt} 
\begin{tabular}{@{}cccccc@{}}
\toprule
Data & kHz & Train (h) & Valid (h) & Test (h) & Speech \\
\midrule
WSJ & 16 & 81.5 & 1.1 & 0.7 & read \\
TL & 16 & 452 & 1.6 & 2.6 & oratory \\
CV & 48 & 693 & 27.1 & 25.8 & read \\
LS & 16 & 960 & 5.1+5.4 & 5.4+5.4 & read \\
SB+FSH & 8 & 300+2k & 6.3 & 1.7+2.1 & conversational \\
RV & 16 & 5k & 14.4 & 18.8+19.5+37.2 & diverse \\
\bottomrule
\end{tabular}
\end{center}
\end{table}

\begin{table}[t!]
\caption{Statistics on datasets: mean sample duration (in seconds) and mean sample transcription length (in words). \label{tab:datadur}}
\begin{center}
\resizebox{\linewidth}{!}{
\setlength\tabcolsep{6pt} 
\begin{tabular}{@{}cccccccc@{}}
\toprule
Data & Train $\mu\pm\sigma$ (s) & Valid $\mu\pm\sigma$ (s) & Test $\mu\pm\sigma$ (s) & Train $\mu\pm\sigma$ (wrd) & Valid $\mu\pm\sigma$ (wrd) & Test $\mu\pm\sigma$ (wrd) \\
\midrule
WSJ & $7.8 \pm 2.9$ & $7.8 \pm 2.9$ & $7.6 \pm 2.5$ & $17 \pm 7$ & $16\pm 7$ & $17\pm6$ \\
TL & $6 \pm 3$ & $11.3 \pm 5.7$ & $8.1 \pm 4.3$ & $17 \pm 10$ & $35\pm 20$ & $24\pm 15$ \\
CV & $5.7 \pm 1.6$ & $6.1 \pm 1.8$ & $5.8 \pm 2.6$ & $10 \pm 3$ & $10\pm 3$ & $9\pm 3$ \\
LS & $12.3 \pm 3.8$ & $6.8 \pm 4.5$ & $7 \pm 4.8$ & $33 \pm 12$ & $19\pm 13$ & $19\pm13$ \\
SB+FSH & $3.7 \pm 3.2$ & $4 \pm 3.1$ & $2.1 \pm 1.7$ & $11 \pm 12$ & $12\pm 12$ & $8 \pm 8$ \\
RV & $8.5 \pm 1.9$ & $11.6 \pm 2.8$ & $11.6 \pm 2.7$ & $21 \pm 10$ & $25 \pm 13$ & $29 \pm 12$ \\
\bottomrule
\end{tabular}
}
\end{center}
\vspace{-0.4cm}
\end{table}

\subsection{Unifying Audio}

The datasets used in our work have different sample rate and varied input lengths as shown in Table~\ref{tab:data} and \ref{tab:datadur}. Since we require the same set of filterbanks for joint training across all datasets, we upsample/downsample each dataset to 16kHz and use this setup for training both baseline models on individual datasets as well as joint models. For all experiments we compute 80 log-mel spectrogram features for a 25ms sliding window, strided by 10ms. All features are normalized to have zero mean and unit variance per input sequence before feeding into the neural network.

\begin{figure}[ht!]
\centering
    \begin{tikzpicture}
    \begin{groupplot}[group style = {group size = 1 by 2, horizontal sep = 0pt}, width =4.7cm, height = 6.0cm]
        \nextgroupplot[
        width = 14cm, height = 5cm,
        ytick={-3, -1, 1},
        xtick={0,20,40,60,80},
        ylabel={\small Energy},
        xlabel={\small Filterbank Index},
        every x tick scale label/.style={at={(xticklabel cs:0.9)},anchor=south west},
        legend style={draw=none,at={(1,0.4)},legend columns=2,font=\scriptsize}
        ]
            \legend{RT03S 16kHz, TL 16kHz, RT03S 8kHz, TL 8kHz}
            \addplot[red] coordinates { (0, -2.272122324060747) (1, -1.5593899399168731) (2, -1.8742556336681877) (3, -1.6380375132066731) (4, -1.4217320772680317) (5, -1.6666477606543955) (6, -0.9654331465944616) (7, -1.0325240219967688) (8, -0.9112988478168872) (9, -0.4773573971625552) (10, -0.6030902644709676) (11, -0.5674086779342398) (12, -0.27349666150395935) (13, -0.3829198670360332) (14, -0.2624646575294775) (15, -0.22471691382926115) (16, 0.010548845895755837) (17, -0.07049900488056347) (18, -0.00478090362090989) (19, 0.027089921786168435) (20, 0.0447890213344707) (21, 0.05880014393248267) (22, 0.07861880766546839) (23, 0.1055115437172495) (24, 0.07569990336723526) (25, 0.08202763819854735) (26, 0.07826930082300602) (27, 0.07248454950735851) (28, 0.06557397952560548) (29, 0.06728858005768758) (30, 0.08052582700946809) (31, 0.05405787316549937) (32, 0.051007172186065146) (33, 0.049195153553184566) (34, 0.04756840297483067) (35, 0.06013180403858702) (36, 0.09686493041666569) (37, 0.03793217754589079) (38, 0.04007602591308596) (39, 0.130250555302804) (40, 0.06676436198693593) (41, 0.1042436874996297) (42, 0.14727893841462622) (43, 0.12563110738516142) (44, 0.20114940566560968) (45, 0.1881981395243601) (46, 0.23431307770050644) (47, 0.2763213087299679) (48, 0.27786781325704) (49, 0.3300513708977514) (50, 0.3663243793540902) (51, 0.37115096837738015) (52, 0.3946274030859749) (53, 0.4069314082808839) (54, 0.4148655361079874) (55, 0.4195036937212738) (56, 0.4215381479295952) (57, 0.42843067486025205) (58, 0.4415290348235379) (59, 0.45653653709840747) (60, 0.47531421903398485) (61, 0.504020961234861) (62, 0.5310767728999387) (63, 0.5235082960672192) (64, 0.5471075230266611) (65, 0.5340336230670327) (66, 0.5386652961617886) (67, 0.5126859562362162) (68, 0.5310987597514334) (69, 0.5113476670015156) (70, 0.508559735432537) (71, 0.5107433191531645) (72, 0.5098494560769454) (73, 0.5037973634052907) (74, 0.4864256901758033) (75, 0.4532047591853765) (76, 0.40731328863825594) (77, 0.3518140225172477) (78, 0.1901775978369989) 
            };
            \addplot[blue] coordinates { (0, -2.778392074966106) (1, -1.6698775225678595) (2, -1.7359558747771022) (3, -1.391574967201075) (4, -0.832454299661932) (5, -1.1834370836828583) (6, -0.25315923887221514) (7, -0.6611874173662788) (8, -0.49236109955507257) (9, -0.12075208221743342) (10, -0.38027083371119896) (11, -0.3596631150457081) (12, -0.004499514822331822) (13, -0.23901323305890304) (14, -0.13858613501739275) (15, -0.14923044569930366) (16, 0.05813485742563515) (17, -0.09604239971783689) (18, 0.0059227706103226465) (19, 0.05078739678749967) (20, 0.06024740754629015) (21, 0.03979924194717746) (22, 0.035469048767974104) (23, 0.05966119638635696) (24, 0.027568262683037234) (25, 0.036664437090194325) (26, 0.02162403801428151) (27, -0.0032080747727202064) (28, -0.030749686609251915) (29, -0.04650971696215062) (30, -0.04779613081033299) (31, -0.0999158846253589) (32, -0.11442970325162043) (33, -0.12843577521591382) (34, -0.1408460604088698) (35, -0.13206986358700953) (36, -0.08178890268092875) (37, -0.14809968291711129) (38, -0.13956133830266493) (39, -0.03375224313292035) (40, -0.1290746373385865) (41, -0.06536512680578983) (42, 0.0027343193434370334) (43, -0.022886718741563714) (44, 0.08581056182341193) (45, 0.06393448702558317) (46, 0.1209681945625908) (47, 0.17484146222960464) (48, 0.15918866522260555) (49, 0.21582874770586968) (50, 0.2527251951061318) (51, 0.2509255917686873) (52, 0.2750261570976736) (53, 0.2897210760801405) (54, 0.29580346115383205) (55, 0.30238762601691604) (56, 0.31464941898907195) (57, 0.3371592274658575) (58, 0.36969703724220726) (59, 0.413126133329381) (60, 0.4530234697458007) (61, 0.49086238555792305) (62, 0.5258523589301757) (63, 0.5173142752135054) (64, 0.5536796528661713) (65, 0.5266630576679794) (66, 0.5284266570821161) (67, 0.491418353123482) (68, 0.5192174498839129) (69, 0.4991327268326681) (70, 0.4962378705867401) (71, 0.48635533475183323) (72, 0.4728465328165019) (73, 0.45194982567549763) (74, 0.4359271921689742) (75, 0.4245451606822816) (76, 0.4145607648384462) (77, 0.41839355975685594) (78, 0.4299407588054477) (79, 0.394193477695221)
            };
            \addplot[red,dash pattern=on 2pt off 2pt] coordinates { (0, -1.5543147195868041) (1, -1.4716271241126786) (2, -1.412085065052064) (3, -1.1041557636758412) (4, -0.9421256748472586) (5, -0.6662334842180602) (6, -0.4417822136507099) (7, -0.27670334052754425) (8, -0.17902581278136626) (9, -0.10696516838668735) (10, -0.006219058764304631) (11, 0.110468587118521) (12, 0.20422983972048098) (13, 0.25838728599473937) (14, 0.28558266513144565) (15, 0.3128634318510568) (16, 0.33740198873365235) (17, 0.3672973936107101) (18, 0.40777016963500085) (19, 0.3324267840290906) (20, 0.32001295403481056) (21, 0.3960638303000742) (22, 0.28894193129248746) (23, 0.3565841381424489) (24, 0.2977024169428026) (25, 0.3505586932475293) (26, 0.29969890872918603) (27, 0.3400592717663235) (28, 0.33411995178561393) (29, 0.3386535054858393) (30, 0.3595934570019639) (31, 0.3796103359316656) (32, 0.40816774432427033) (33, 0.44078428765039945) (34, 0.4776208303796013) (35, 0.5147963554114899) (36, 0.5530909459422891) (37, 0.5879398117995079) (38, 0.6460573768033092) (39, 0.6437374878831208) (40, 0.6746445665944407) (41, 0.6685261561711169) (42, 0.6925363037385043) (43, 0.6857745376044166) (44, 0.7086360460287486) (45, 0.736290442939181) (46, 0.7654735861454678) (47, 0.7877305939328706) (48, 0.7942500579485987) (49, 0.795127455223731) (50, 0.7920962934980421) (51, 0.7717343096523782) (52, 0.7764061359139279) (53, 0.7678184492045664) (54, 0.765477594400561) (55, 0.7575906245766517) (56, 0.7297376614128882) (57, 0.6770049670874263) (58, 0.603921680533081) (59, 0.27574658347692865) (60, -0.5731804957765916) (61, -0.8410016281174308) (62, -0.8521449790392188) (63, -0.8631770701531786) (64, -0.8703926512643355) (65, -0.8781282974970606) (66, -0.8852180999862412) (67, -0.8900364176875601) (68, -0.8919425485021133) (69, -0.8930589034440075) (70, -0.8921572324873738) (71, -0.8842533712633593) (72, -0.882361469564388) (73, -0.874813070192128) (74, -0.8670657187113608) (75, -0.8559642174303976) (76, -0.8482637931792505) (77, -0.8364386387278407) (78, -0.8239160980052566) (79, -0.8119943001308196) 
            };
            \addplot[blue,dash pattern=on 2pt off 2pt] coordinates { (0, -2.3135316642833126) (1, -1.9315972597321807) (2, -1.6729236666322311) (3, -0.9302120196690077) (4, -0.7454661188314317) (5, -0.5784005197660886) (6, -0.45410286214235934) (7, -0.3646919727133901) (8, -0.31122769501083947) (9, -0.26736736244491577) (10, -0.21401326354034295) (11, -0.17273147660288415) (12, -0.14410083025811585) (13, -0.0956664880974879) (14, -0.043723080313626046) (15, -0.032074326940544004) (16, -0.04533255616927838) (17, -0.028284558911952168) (18, 0.03449082774528806) (19, -0.0796788541605633) (20, -0.11737783836514804) (21, -0.03364873183722398) (22, -0.2165931828945416) (23, -0.14068683072948918) (24, -0.24426223049068344) (25, -0.18284977380638018) (26, -0.27138184341686616) (27, -0.2123940457862434) (28, -0.20510982840889652) (29, -0.20863649274766088) (30, -0.18852134449717653) (31, -0.14347312428715467) (32, -0.09155908529646721) (33, -0.03761494998020529) (34, 0.011226956405543637) (35, 0.059480420259743115) (36, 0.09638031432306948) (37, 0.12556387132818678) (38, 0.19646431090137043) (39, 0.18155059922720074) (40, 0.2218339564361211) (41, 0.20768216148512078) (42, 0.2532161279721896) (43, 0.2619912665381704) (44, 0.3223097849231697) (45, 0.3885657625377502) (46, 0.4321071970907723) (47, 0.46361699425724145) (48, 0.47735022764973173) (49, 0.4670876044631664) (50, 0.4557001444969028) (51, 0.4271506643340026) (52, 0.4436904529803211) (53, 0.42652902488662003) (54, 0.40691552550410826) (55, 0.38083647662958064) (56, 0.36070942679283874) (57, 0.34516642873566367) (58, 0.3562576141470549) (59, 0.3372381931142149) (60, 0.31409374650445887) (61, 0.2530418438659231) (62, 0.2280053551331226) (63, 0.20631670182677556) (64, 0.20212898777889718) (65, 0.20960442394502374) (66, 0.22459482201038808) (67, 0.2620215164902449) (68, 0.28665930025737185) (69, 0.2814870287785782) (70, 0.27948963142726985) (71, 0.30327549414899524) (72, 0.3129277784304177) (73, 0.3149064606199842) (74, 0.2977600492627996) (75, 0.26910850745744813) (76, 0.24698523299193212) (77, 0.25741846402904084) (78, 0.22004775244171618) (79, -0.39174955380089665) 
            };
    \end{groupplot}
\end{tikzpicture}
\caption{Distribution of the mean normalized energy of 80 filterbanks on some public validation sets we used, for 16kHz audio (dashed) and 8kHz audio (solid).}
\label{fig:features}
\vspace{-0.2cm}
\end{figure}
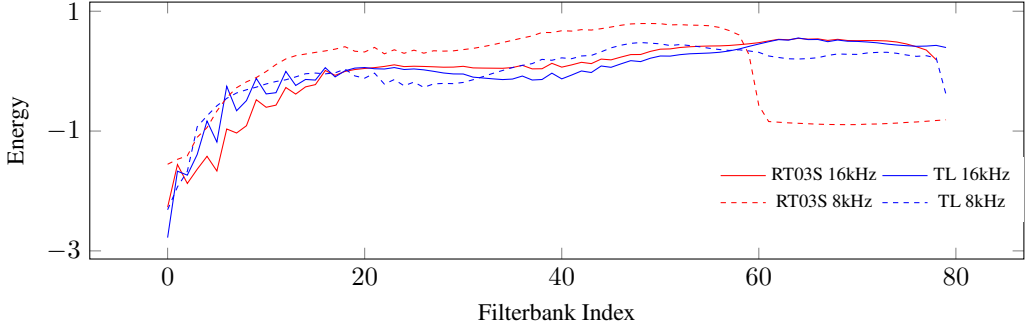

On SB+FSH individual baseline, we span the log-mel filterbanks up to only 4kHz (unlike 8kHz for all other training setups) as any spectrogram features beyond 4kHz cannot be determined accurately~\citep{shannon1949comm}. This can also be seen in Figure~\ref{fig:features} which plots the distribution of mean normalized energy of filterbanks for different datasets with audio sampled at 16kHz and filterbanks span from 0-8kHz.

\subsection{Language Model}
We train a $n$-gram LM using KenLM toolkit~\citep{heafield2011kenlm} and a Transformer LM similar to~\citep{synnaeve2019end} for each dataset independently using their in-domain LM training corpus. Specifically, we use the training transcriptions as LM corpus for domains like SB+FSH and RV; while for TL, both training transcriptions and the original LM corpus are combined together to train its LM. All the Transformer LMs share the similar architecture as~\citep{baevski2018adaptive}'s Google Billion Words model: we use 8 attention heads; 8 (WSJ, CV and SB+FSH), 16 (TL) or 20 (LS) decoder layers with embedding, input and output dimensions of 512 (CV), 1024 (WSJ and SB+FHS) or 1280 (TL and LS); feed-forward layer dimension is set to 1024 ( CV), 2048 (WSJ and SB+FHS) or 6144 (TL and LS); dropout is 0.3 (WSJ, CV and SB+FHS), 0.15 (TL) or 0.1 (LS). Number of decoder layers, embedding dimensions as well as dropout were tuned on each dataset depending on the amount of training data. 

We also train a 4-gram and a Transformer LMs on Common Crawl (CC) data~\citep{wenzek2020ccnet}. Before any training we perform the following text normalization for CC data: splitting paragraphs into separate sentences, punctuation removal, mapping of common abbreviations, converting latin and roman numbers into the text. We keep a dictionary of 200k most-common words. For 4-gram training we prune all 3,4-grams appearing once and use only 10\% of the CC data. The transformer LM also follows the~\citep{baevski2018adaptive}'s Google Billion Words model and trained on all CC data: we use 8 attention heads and 16 decoder layers with embedding, input and output dimensions 1024 and feed-forward layer dimension 4096, dropout is set to 0.1. The perplexity of all LMs are shown in Table~\ref{tab:lm}.

 \begin{table}[h!]
\caption{
Perplexity (including out-of-vocabulary words) of word-level LMs. We use 4-gram LMs for WSJ, LS, SB+FSH, and 5-gram LMs for TL, CV.
\label{tab:lm}}
\begin{center}
\setlength\tabcolsep{6pt} 
\begin{tabular}{@{}ccccccccc@{}}
\toprule
\multirow{2}{*}{Data/Vocab} & \multicolumn{2}{c}{in-dom. $n$-gram} &\multicolumn{2}{c}{in-dom. Transf.} &\multicolumn{2}{c}{CC $4$-gram} &\multicolumn{2}{c}{CC Transf.} \\
\cmidrule(lr){2-3} \cmidrule(lr){4-5} \cmidrule(lr){6-7} \cmidrule(lr){8-9}
  & Valid & Test & Valid & Test & Valid & Test & Valid & Test \\
\midrule
WSJ/162K & 159 & 134 & 84 & 69 & 297 & 285 & 119 & 116 \\
TL/200k & 119 & 149 & 74 & 79 & 142 & 136 & 82 & 80 \\
CV/168K & 359 & 329 & 181 & 188 & 213 & 157 & 119 & 100\\
LS/200K & 155/147 & 164/154 & 48/50 & 52/50 & 258/258 & 244/249 & 133/137 & 139/137 \\
SB+FSH/64K & 124 & 114/112 & 50 & 55/67 & 221 & 199/153 & 70 & 67/83 \\
RV/200K & 158 & 146 & - & - & 249 & 204 & - & - \\
\bottomrule
\end{tabular}
\end{center}
\vspace{-0.2cm}
\end{table}

To integrate LMs with AMs, we use one-pass beam-search decoder from the wav2letter++~\citep{collobert2016wav2letter} (lexicon-based with a $n$-gram LM) and an additional second-pass rescoring with a Transformer LM following~\citep{synnaeve2019end}.

\begin{table*}[t!]
\caption{
WER of models evaluated on all datasets (downsampled to 16kHz) with a greedy decoding and {\it no LM} (top row), {\it with in-domain n-gram LM} beam-search decoding (middle row) and with additional second-pass rescoring {\it by in-domain Transformer LM} (below row). Joint models are also decoded with CC LM with either a single-pass (top row) or a two-pass (bottom row) decoding. State-of-the-art (SOTA) models are given from WSJ~\citep{hadian2018end}, TED-LIUM~\citep{zhou2020rwth}, LibriSpeech~\citep{gulati2020conformer}, SwitchBoard \& Fisher~\citep{han2017capio}. The SOTA models are all decoded with in-domain LMs. The average is computed as average of averages for LibriSpeech's validations/tests, and SwitchBoard's tests (SB, CH) sets, so as not to weight them more heavily.
\label{tab:all}}
\begin{center}
\begin{small}
\setlength\tabcolsep{3pt} 
\npdecimalsign{.}
\nprounddigits{1}
\npnoroundexp
\resizebox{\linewidth}{!}{
\begin{tabular}{cn{2}{1}n{2}{1}n{2}{1}n{2}{1}n{2}{1}n{2}{1}n{2}{1}n{2}{1}n{2}{1}n{2}{1}n{2}{1}n{2}{1}n{2}{1}n{2}{1}n{2}{1}}
\toprule
\multirow{2}{*}{Train} & \multicolumn{2}{c}{WSJ} & \multicolumn{2}{c}{TL} & \multicolumn{2}{c}{CV} & \multicolumn{4}{c}{LS} & \multicolumn{3}{c}{SB+FSH} &  \multicolumn{2}{c}{average} \\
\cmidrule(lr){2-3} \cmidrule(lr){4-5} \cmidrule(lr){6-7} \cmidrule(lr){8-11} \cmidrule(lr){12-14} \cmidrule(lr){15-16}
 & \begin{footnotesize}\texttt{nov93}\end{footnotesize} & \begin{footnotesize}\texttt{nov92}\end{footnotesize} &
 \begin{footnotesize}\texttt{valid}\end{footnotesize} & \begin{footnotesize}\texttt{test}\end{footnotesize} &
 \begin{footnotesize}\texttt{valid}\end{footnotesize} & \begin{footnotesize}\texttt{test}\end{footnotesize} &
 \begin{footnotesize}\texttt{dev-c}\end{footnotesize} & \begin{footnotesize}\texttt{test-c}\end{footnotesize} & \begin{footnotesize}\texttt{dev-o}\end{footnotesize} & \begin{footnotesize}\texttt{test-o}\end{footnotesize} & \begin{footnotesize}\texttt{RT03S}\end{footnotesize} & \begin{footnotesize}\texttt{SB}\end{footnotesize} & \begin{footnotesize}\texttt{CH}\end{footnotesize} & \begin{footnotesize}\texttt{valid}\end{footnotesize} & \begin{footnotesize}\texttt{test}\end{footnotesize} \\
\midrule
SOTA &  & 2.8 & 5.1 & 5.6 &  &  &  & 1.9  & & 3.9 & 8.0 & 5.0 & 9.1 &  &  \\
\midrule
\midrule
\multirow{3}{*}{WSJ} & \cellcolor{gray!20}13.5 & \cellcolor{gray!20}11.7 & 48.79 & 42.1 & 72.21 & 78.02 & 32.43 & 32.66 & 53.81 & 53.99 & 78.3 & 69.6 & 84.6 & 51.17 & 50.44 \\
 & \cellcolor{gray!20}~~7.3 & \cellcolor{gray!20}~~5.3 & 35.19 & 28.74 & 54.99 & 62.03 & 17.72 & 18.24 & 37.43 & 38.82 & 68 & 58.2 & 78.3 & 38.61 & 38.58 \\
 & \cellcolor{gray!20}~~5.7 & \cellcolor{gray!20}~~4.1 & 33.85 & 26.85 & 53.49 & 60.71 & 14.64 & 15.21 & 35.23 & 36.67 & 66.8 & 57.4 & 77.5 & 36.94 & 37.0  \\  
\midrule
\multirow{3}{*}{TL} & 12.0 & 9.89 & \cellcolor{gray!20}10.2 & \cellcolor{gray!20}~~7.9 & 32.16 & 36.7 & 11.82 & 12.42 & 21.64 & 22.6 & 32.1 & 23.6 & 32.4 & 20.64 & 19.99 \\
 & 7.59 & 5.76 & \cellcolor{gray!20}~~7.9 & \cellcolor{gray!20}~~6.5 & 23.93 & 28.13 & 7.54 & 8.41 & 15.33 & 16.35 & 27.3 & 19.3 & 27.7 & 15.62 & 15.26 \\
 & 6.28 & 5.14 & \cellcolor{gray!20}~~7.4 & \cellcolor{gray!20}~~6.2 & 22.63 & 27.31 & 5.89 & 6.94 & 13.11 & 14.14 & 26.8 & 19 & 27.3 & 14.53 & 14.46 \\
\midrule
\multirow{3}{*}{CV} & 12.84 & 9.7 & 68.54 & 46.99 & \cellcolor{gray!20}12.0 & \cellcolor{gray!20}15.4 & 35.51 & 36.3 & 37.03 & 39.08 & 49.2 & 46.9 & 45.7 & 35.76 & 31.22 \\
 & 5.99 & 3.51 & 56.57 & 34.16 & \cellcolor{gray!20}10.1 & \cellcolor{gray!20}13.0 & 23.75 & 25.55 & 26.6 & 29.07 & 39.2 & 36.2 & 36.6 & 27.41 & 22.87 \\
 & 5.45 & 3.24 & 55.33 & 32.44 & \cellcolor{gray!20}10.0 & \cellcolor{gray!20}12.8 & 21.94 & 23.72 & 24.36 & 26.91 & 37.8 & 35.0 & 35.5 & 26.34 & 21.82 \\
\midrule
\multirow{3}{*}{LS-960} & 9.96 & 7.85 & 13.63 & 12.97 & 25.8 & 29.9 & \cellcolor{gray!20}~~2.6 & \cellcolor{gray!20}~~2.7 & \cellcolor{gray!20}~~7.0 & \cellcolor{gray!20}~~6.8 & 36.6 & 27.6 & 35 & 18.16 & 17.35 \\
 & 4.72 & 3.49 & 8.96 & 9.72 & 18.56 & 22.27 & \cellcolor{gray!20}~~2.0 & \cellcolor{gray!20}~~2.5 & \cellcolor{gray!20}~~5.2 & \cellcolor{gray!20}~~5.5 & 28.3 & 20.0 & 27.8 & 12.83 & 12.68 \\
 & 3.92 & 2.92 & 8.45 & 8.83 & 17.63 & 21.62 & \cellcolor{gray!20}~~1.5 & \cellcolor{gray!20}~~2.0 & \cellcolor{gray!20}~~4.2 & \cellcolor{gray!20}~~4.5 & 27.7 & 20 & 27.1 & 12.12 & 12.04 \\
\midrule
\multirow{3}{*}{SB+FSH} & 10.91 & 9.46 & 15.05 & 12.38 & 49.39 & 50.66 & 13.96 & 14.41 & 27.36 & 28.6 & \cellcolor{gray!20}12.0 & \cellcolor{gray!20}~~6.9 & \cellcolor{gray!20}11.4 & 21.60 & 20.63 \\
 & 5.09 & 4.0 & 9.34 & 8.89 & 40.42 & 41.8 & 7.46 & 8.19 & 18.75 & 20.41 & \cellcolor{gray!20}10.4 & \cellcolor{gray!20}~~6.5 & \cellcolor{gray!20}10.3 & 15.67 & 15.44 \\
 & 4.23 & 3.38 & 8.63 & 8.04 & 38.86 & 40.52 & 5.38 & 6.21 & 16.3 & 18.14 & \cellcolor{gray!20}10.4 & \cellcolor{gray!20}~~6.5 & \cellcolor{gray!20}10.3 & 14.59 & 14.46 \\
\midrule
\midrule
\multirow{3}{*}{Joint} & 3.0 & 2.04 & 6.08 & 5.69 & 11.13 & 13.22 & 2.49 & 2.52 & 6 & 5.9 & 10.7 & 5.8 & 9.7 & 7.03 & 6.59 \\
 & 1.97 & 1.4 & 5.37 & 5.46 & 9.39 & 11.08 & 1.82 & 2.31 & 4.49 & 4.8 & 9.4 & 5.4 & 8.6 & 5.86 & 5.71  \\
& 1.7 & 1.3 & 5.0 & 4.7 & 9.2 & 10.9 & 1.4 & 2.0 & 3.7 & 4.1 & 9.4 & 5.4 & 8.6 & 5.6 & 5.4 \\
\midrule
 \multirow{2}{*}{Joint CC} & 2.81 & 2.0 & 5.58 & 5.08 & 8.11 & 9.38 & 2.9 & 2.94 & 5.27 & 5.3 & 9.6 & 5.4 & 8.7 & 6.037 & 5.53 \\
 & 2.78 & 1.89 & 5.34 & 4.77 & 8.0 & 9.29 & 2.87 & 2.92 & 5.23 & 5.27 & 8.9 & 5.4 & 8.7 & 5.81 & 5.41 \\
\midrule
\midrule
\multirow{3}{*}{Joint+noise} & 2.99 & 2.18 & 6.30 & 5.81 & 11.246 & 13.33 & 2.48 & 2.58 & 6.051 & 5.98 & 10.8 & 6.3 & 10.3 & 7.12 & 6.78 \\
& 1.99 & 1.49 & 5.48 & 5.247 & 9.41 & 11.16 & 1.83 & 2.32 & 4.549 & 4.87 & 9.4 & 5.8 & 9.3 & 5.89 & 5.81 \\
& 1.88 & 1.346 & 5.13 & 4.88 & 9.32 & 11.02 & 1.43 & 2.03 & 3.66 & 4.16 & 9.4 & 5.8 & 9.3 & 5.655 & 5.58 \\
\midrule
 Joint+noise CC  & 2.83 & 1.98 & 5.74 & 5.18 & 8.18 & 9.57 & 2.92 & 3.00 & 5.24 & 5.30 & 9.5 & 5.6 & 9 & 6.07 & 5.64 \\
\bottomrule
\end{tabular}
} 
\npnoround
\end{small}
\end{center}
\vspace{-0.4cm}
\end{table*}

\subsection{Baselines and Joint Acoustic Model}
\subsubsection{Acoustic Model (AM)}
All models are trained with Connectionist Temporal Classification~\citep{graves2006connectionist} and 
the network architecture follows~\citep{synnaeve2019end}: the encoder of our AMs is composed of a convolutional frontend (1-D convolution with kernel-width~7 and stride~3 followed by GLU activation) followed by sinusoidal positional embedding and 36 4-heads Transformer blocks~\citep{vaswani2017attention}. To speed up training we don't use any relative positional embedding inside Transformer blocks. The self-attention dimension is $768$ and the feed-forward network (FFN) dimension is $3072$ in each Transformer block.
The output of the encoder is followed by a linear layer to the output classes. 

We use dropout after the convolution layer. For all Transformer layers, we use dropout on the self-attention and on the FFN, and layer drop~\citep{fan2019reducing}, dropping entire layers at the FFN level. Dropout and layer dropout values are tuned for each model separately.
Token set for all AMs consists of 26 English alphabet letters, augmented with the apostrophe and a word boundary token. The popular approach with word-pieces as tokens set we found to be not suited as intersection between word-pieces constructed on every training set less than 50\%. Thus the question what word-pieces set should be used for the joint model is still open. 
SpecAugment~\citep{park2019specaug} is used for data augmentation in training: there are two frequency masks, and ten time masks with maximum time mask ratio of $p=0.05$; frequency and time mask parameters are tuned separately for each model; time warping is not used. In the joint model, the maximum frequency bands masked by one frequency mask is 30, and the maximum frames masked by the time mask is 30, too.
We use the Adagrad optimizer~\citep{duchi2011adaptive} and decay learning rate by a factor of 2 each time the WER reaches a plateau on the validation sets. All experiments are implemented within flashlight\footnote{\url{https://github.com/flashlight/flashlight}} and wav2letter++~\citep{pratap2018wav2letter}. All models are trained with dynamic batching (effective average batch size is 240s per GPU) and mixed-precision computations on 16 GPUs (Volta 32GB) for 1-3 days for single dataset baselines and 14 days for joint training.

 \begin{table}[t!]
\caption{
WER comparison on CHiME-6 dev and eval sets. Both HMM model results are obtained by decoding with a 3-gram LM, the second HMM model being the single best AM in the CHiME 2020 challenge on this track. All other results are with {\it a greedy decoding} without an LM. Both HMM models and the RNN-T models are trained on CHiME-6. All other models correspond to the ones in Table~\ref{tab:all} and do not use CHiME-6 for training (direct transfer).
\label{tab:chime}}
\begin{center}
\begin{tabular}{cccc}
\toprule
Model & Train & Dev & Eval \\
\midrule
Official HMM~\citep{watanabe2020chime} & CHiME-6 &   51.8 & 51.3 \\
HMM~\citep{medennikov2020stc} & CHiME-6 & 36.9 & 38.6 \\
RNN-T~\citep{andrusenko2020towards} & CHiME-6 & 49.0 & - \\
\midrule
\multirow{7}{*}{Transformer (ours)} & WSJ & 97.5 & 95.8 \\
 & TL & 61.7 & 70.5 \\
 & CV & 81.2 & 77.9 \\
 & LS-960 & 74.4 & 79.6 \\
 & SB+FSH & 55.9 & 67.0 \\
\cline{2-4} 
 & Joint & 44.9 & 56.9 \\
 & Joint+noise & 41.5 & 51.7 \\
\bottomrule
\end{tabular}
\vspace{-0.2cm}
\end{center}
\end{table}

\subsubsection{Joint Model}
We adopt the same AM architecture described above but with less regularization when training on the combination of all the datasets. We weight each sample equally, i.e. each sample from each dataset is fed into the model once in each epoch. 

\subsubsection{Joint+noise Model}
To further improve the robustness of our joint model, we fine-tune the model by using  data augmentation on the training data. In our work, we use two popular audio data augmentation procedures: additive noise and reverberation. For additive noise, we randomly sample a audio clip from Audioset~\citep{gemmeke2017audio} database. For each sample, we randomly sample an value between a chosen min SNR and max SNR values, and scale the noise accordingly before adding it to the input signal. Reverberation is done by convolving the input signal with a randomly sampled RIR. Similar to~\citep{balam2020improving}, we consider real and simulated room impulse responses (RIRs) from OpenSLR~\citep{ko2017study} and BUT ReverDB~\citep{szoke2019building}. We use 0 and 40 as min and max values for SNR and the probability of applying additive noise, reverberation augmentation is set to 0.4 and 0.2 respectively. We have chosen these settings based on the average performance of joint+noise model on all validation sets.

\subsubsection{Joint and Joint+noise Models Fine-tuning on RV}

We fine-tune our best joint model and joint-noise model on small parts of RV training data, to see how they transfer to real-world noisy data. The subsets are randomly selected to have length 10 minutes, 1 hour, 10 hours and 100 hours. 
To reach the best performance on different amount of training data, we conduct a grid search over the following parameters: learning rate in (0.001, 0.005, 0.01), warm-up updates in (1000, 2000, 4000, 6000), and learning rate decreasing scheduler. The optimal learning rate and warm-up updates are 0.005 and 2000 for all configurations. Learning rate decreases every (20, 200, 2000, 5000) epochs for (10 min, 1hr, 10 hrs and 100hrs) settings. We didn't use any additive noise or reverberation in fine-tuning on RV. All the experiment results are listed in Table~\ref{tab:outdomain_ft}.

 \begin{table}[t!]
\caption{
Direct transfer. WER comparison with {\it a greedy decoding} and with {\it a $5$-gram in-domain LM} and {\it the 4-gram CC LM} beam-search decoding on RV validation and test data from videos. Except for the in-domain ``RV'' training and for models with ``+finetune'' (in-domain finetuning), all other models correspond to models in Table~\ref{tab:all}. 
\label{tab:outdomain}}
\begin{center}
\setlength\tabcolsep{6pt} 
\begin{tabular}{@{}cccccc@{}}
\toprule
\multirow{2}{*}{Train} & \multirow{2}{*}{LM} & \multirow{2}{*}{Valid} & \multicolumn{3}{c}{Test} \\
\cmidrule(lr){4-6}
 & & & clean & noisy & extreme \\
\midrule
 \multirow{3}{*}{RV (5000h)} & - & 26.3  & 14.6  & 19.1 & 27.9  \\  
 & in-dom. & 22.9 & 12.3 & 16.2 & 24.1 \\
 & CC & 22.9 & 12.1 & 16.1 & 24.2\\
\midrule
\midrule
\multirow{3}{*}{WSJ (81.5h)} & - & 78.9  & 66.3 & 73.1 & 80.2  \\
 & in-dom. & 68.4 & 54.0 & 61.9 & 69.8 \\
 & CC & 69.3 & 54.1 & 62.5 & 70.9  \\
\midrule
\multirow{3}{*}{TL (452h)} & - & 42.5 & 24.1 & 32.0 & 44.1  \\
& in-dom. & 36.0 & 19.1 & 26.3 & 37.1 \\
 & CC & 36.1  & 19.0 & 26.3 & 37.5  \\
\midrule
\multirow{3}{*}{CV (693h)} & - & 73.2 & 65.4 & 70.5 & 73.5 \\
 & in-dom. & 62.5 & 51.4 & 57.9 & 63.2  \\
 & CC & 63.0 & 51.8 & 58.5 & 64.0  \\
\midrule
\multirow{3}{*}{LS-960 (960h)} & - & 48.0 & 28.5 & 37.6 & 50.3  \\
 & in-dom. & 38.2 & 21.5 & 29.5 & 40.2 \\
 & CC & 38.9 & 21.7 & 29.9 & 41.2  \\
\midrule
\multirow{3}{*}{SB+FSH (2300h)} & - & 43.7 & 28.8 & 33.8 & 44.7 \\
 & in-dom. & 40.3 & 24.6 & 29.9 & 41.3\\
 & CC & 40.4 & 24.5 & 30.0 & 41.5  \\
\midrule
\midrule
\multirow{3}{*}{Joint (4500h)} & - & 39.8 & 16.0 & 22.4 & 32.9  \\
& in-dom. & 28.3 & 13.5 & 19.3 & 29.6\\
 & CC & 28.7 & 13.6 & 19.5 & 30.0  \\
 \midrule
 \multirow{3}{*}{Joint+noise (4500h)} & - & 30.7 & 15.6 & 21.5 & 32.7  \\
& in-dom. & 27.3 & 13.1 & 18.2 & 29.2  \\
 & CC & 27.6 & 13.1 & 18.5 & 29.7 \\
\bottomrule
\end{tabular}
\vspace{-0.85cm}
\end{center}
\end{table}

\subsection {AM Transfer} 

In general, an AM trained in isolation on a single dataset performs poorly on other datasets, as shown in Table~\ref{tab:all}. The model trained on WSJ performs the worst (part of the reason could be the smaller amount of training data) for transfer, while other models transfer very well to WSJ. All models transfer poorly to CV and the CV model transfers  poorly to other datasets, which may indicate that CV is very different from other benchmarks. From the results on LS, TL and SB+FSH there is a similarity between LS and TL (they transfer the best to each other). There is also a similarity in transfer between SB+FSH and TL benchmarks, however, LS and SB+FSH do not transfer well to each other. When training on all datasets at once, the joint model in Table~\ref{tab:all} performs better or close to a single dataset training. This behaviour compared to results on a single dataset training indicates that i) datasets differ from each other and ii) a robust model scoring well on all these benchmarks exists.

We also test our models from Table~\ref{tab:all} on CHiME-6 dev and eval sets to analyse the noise adaptation, see Table~\ref{tab:chime}. The best single-dataset models are TL and SB+FSH, probably due to the source of their data, like conversational and oratory speech with noisy conditions. Our joint model improves noise adaptation significantly over the best single-dataset models. Finally, the joint+noise model improves further results, stating that additive noise and reverberation augmentation training helps to improve noise robustness.

In Table~\ref{tab:outdomain}, we report results of transfer, of those same models trained on public datasets, to our in-house RV dataset. We also report numbers from a baseline system that is trained in-domain on a corresponding training set of 5000h. As for other benchmarks, single dataset training transfers poorly to in-house data, however, the transfer quality varies a lot, having the best results from the TL model. At the same time our joint model, which performs well on each benchmark, transfers really well, stating that i) public datasets could be the good proxy of training data for real-world ASR, ii) improving average performance on public benchmarks leads to improving performance on real-world noisy data. Our joint+noise model further improves the robustness stating that additive noise and reverberation augmentations improve transfer to real-world noisy data.

In Table~\ref{tab:outdomain_ft}, we report fine-tuning results of our joint and joint+noise models on RV data with different amount of data: 10min, 1h, 10h and 100h. Fine-tuning with only 1h closes the gap with the RV baseline model for both joint and joint+noise models. Fine-tuning of the joint+noise model on extremely small amount of data, 10min and 1h, has substantially better performance than the fine-tuning of the joint model. Thus, pre-training with additive noise and reverberation is important in case of low resource of in-domain data. Fine-tuning with enough in-domain data, 10h or 100h, gives similar results for the joint and joint+noise models, thus we still can adapt to the noise in our in-domain data even without pre-training with additive noise and reverberation. Also fine-tuning of joint or joint+noise models on 10h or 100h data surpasses WER compared to the RV baseline model decoded both with in-domain LM and CC LM.

 \begin{table}[ht!]
\caption{
Effect of in-domain finetuning. WER comparison with {\it a greedy decoding} and with {\it a $5$-gram in-domain LM} and {\it the 4-gram CC LM} beam-search decoding on RV validation and test data from videos. Except for the in-domain ``RV'' training and for models with ``+finetune'', all other models correspond to models in Table~\ref{tab:all}. 
\label{tab:outdomain_ft}}
\begin{center}
\setlength\tabcolsep{6pt} 
\begin{tabular}{@{}cccccc@{}}
\toprule
\multirow{2}{*}{Train} & \multirow{2}{*}{LM} & \multirow{2}{*}{Valid} & \multicolumn{3}{c}{Test} \\
\cmidrule(lr){4-6}
 & & & clean & noisy & extreme \\
\midrule
 \multirow{3}{*}{RV (5000h)} & - & 26.3  & 14.6  & 19.1 & 27.9  \\  
 & in-dom. & 22.9 & 12.3 & 16.2 & 24.1 \\
 & CC & 22.9 & 12.1 & 16.1 & 24.2\\
\midrule
\midrule
\multirow{3}{*}{\shortstack[c]{Joint\\ + finetune RV-10min}} & - & 30.2 & 15.5 & 21.6 & 31.3  \\
 & in-dom. & 26.1 & 12.7 & 18.0 & 27.3 \\
 & CC & 26.4 & 12.7 & 18.1 & 27.6  \\
  \midrule
\multirow{3}{*}{\shortstack[c]{Joint\\ + finetune RV-1h}} & - & 27.3 & 14.0 & 19.2 & 28.3  \\
 & in-dom. & 24.0 & 11.8 & 16.3 & 24.9  \\
 & CC & 24.3 & 11.9 & 16.6 & 25.4 \\
 \midrule
\multirow{3}{*}{\shortstack[c]{Joint\\ + finetune RV-10h}} & - & 25.9 & 12.6 & 17.8 & 27.2  \\
 & in-dom. & 22.9 & 10.8 & 15.3 & 24.2  \\
 & CC & 22.9 & 10.8 & 15.4 & 24.4  \\
 \midrule
 \multirow{3}{*}{\shortstack[c]{Joint\\ + finetune RV-100h}} & - & 25.4 & 12.4 & 17.4 & 26.8  \\
 & in-dom. & 22.5 & 10.6 & 15.0 & 23.8 \\
 & CC & 22.5 & 10.6 & 15.1 & 24.0 \\
 \midrule
 \midrule
\multirow{3}{*}{\shortstack[c]{Joint+noise\\ + finetune RV-10min}} & - & 29.3 & 14.9 & 20.8 & 30.4  \\
 & in-dom. & 25.6 & 12.3 & 17.5 & 26.8  \\
 & CC & 25.9 & 12.3 & 17.6 & 27.1 \\
 \midrule
 \multirow{3}{*}{\shortstack[c]{Joint+noise\\ + finetune RV-1h}} & - & 26.8 & 13.5 & 18.8 & 28.1 \\ 
 & in-dom. & 23.5 & 11.3 & 16.0 & 24.8   \\
 & CC & 24.0 & 11.5 & 16.3 & 25.3 \\
 \midrule
 \multirow{3}{*}{\shortstack[c]{Joint+noise\\ + finetune RV-10h}} & - & 25.9 & 12.7 & 17.9 & 27.3  \\
 & in-dom. & 22.8 & 10.8 & 15.3 & 24.3 \\
 & CC & 22.8 & 11.0 & 15.5 & 24.7 \\
\midrule
 \multirow{3}{*}{\shortstack[c]{Joint+noise\\ + finetune RV-100h}} & - & 25.6 & 12.5 & 17.5 & 26.8  \\
 & in-dom. & 22.6 & 10.6 & 14.9 & 23.8 \\
 & CC & 22.7 & 10.6 & 15.1 & 24.0 \\
\bottomrule
\end{tabular}
\vspace{-0.2cm}
\end{center}
\end{table}

\subsection {Transfer with LM} 
Single-dataset AMs get a boost in WER performance when decoding/rescoring with an in-domain LM, as shown in Table~\ref{tab:all}. These AMs perform however poorly in transfer domain conditions (see Tables~\ref{tab:all} and~\ref{tab:outdomain}). In contrast, the joint models transfers well to in-house RV data, when decoded with an in-domain LM (see Table~\ref{tab:outdomain}). Decoding the joint models with the large generic CC LM leads to WER performance which is overall improved, on both public and in-house~RV datasets, and close to the in-domain LM results.

\subsection{Predictors of transfer}
We performed single variable linear regressions using data from Table~\ref{tab:all}: lines as datapoints, test set score columns as features, and labels being the same models' performance in transfer on the average of RV test clean, noisy, and extreme, from Table~\ref{tab:outdomain}. Across all datasets, and taken over all trained models, the best ``single test set'' predictor for out-of-domain performance on RV data is TL with an $r^2=0.9$ (rejecting the null hypothesis with $p<0.001$), the worst single predictor being CV with $r^2=0.2$ ($p<0.001$). We also performed multivariate regressions using all the test sets from Table~\ref{tab:all} and only the results for the models decoded with $n$-gram LMs. This gives an overspecified problem (more variables: 7, than models: 6), so OLS gives a ``perfect'' (overfitted, $r^2=1$) solution which weights test-other and Callhome {\it a bit negatively}. We repeat this regression with L1 regularization (Lasso, as proxy for L0 norm regularization) with different regularization coefficients. It yields a regression with $r^2\in(0.975, 0.999)$ (albeit with only 6 datapoints) with only TL test set weighted significantly positively and WSJ's nov92 and LS's test-clean weighted at zero. We can conclude that the TL test set is the better predictor, and nov92 and test-clean are the poorest predictors of the performance in transfer on RV of our Transformer-based AMs decoded with $n$-grams. A larger study across AMs and LMs variants should provide a more robust conclusion.

\section{Conclusion}
\label{sec:conclusion}
We studied transfer across five public datasets, as well as transfer to out-of-domain, real-world audio data, for a single AM architecture based on Transformers using non-autoregressive CTC criterion and with $n$-gram and Transformer-based LMs for decoding. We showed that no single validation or test set from public datasets is sufficient to measure transfer to other public datasets or to real-world audio data. Our results suggests that ASR researchers interested in producing transferable AMs should report results on several public datasets, at very least including TED-LIUM (v3). Finally, we provided a recipe for a community-reproducible robust ASR model, which can be trained with a couple of public audio datasets, and language models trained on the Common Crawl dataset.


\bibliography{bibliography.bib}

\begin{thebibliography}{48}
\providecommand{\natexlab}[1]{#1}
\providecommand{\url}[1]{\texttt{#1}}
\expandafter\ifx\csname urlstyle\endcsname\relax
  \providecommand{\doi}[1]{doi: #1}\else
  \providecommand{\doi}{doi: \begingroup \urlstyle{rm}\Url}\fi

\bibitem[Amodei et~al.(2016)]{deepspeech2}
D.~Amodei et~al.
\newblock {Deep Speech 2}: {End-to-End Speech Recognition in English and
  Mandarin}.
\newblock In \emph{ICML}, 2016.

\bibitem[Andrusenko et~al.(2020)Andrusenko, Laptev, and
  Medennikov]{andrusenko2020towards}
A.~Andrusenko, A.~Laptev, and I.~Medennikov.
\newblock Towards a competitive end-to-end speech recognition for chime-6
  dinner party transcription.
\newblock \emph{arXiv preprint arXiv:2004.10799}, 2020.

\bibitem[Ardila et~al.(2020)Ardila, Branson, Davis, Kohler, Meyer, Henretty,
  Morais, Saunders, Tyers, and Weber]{ardila2020common}
R.~Ardila, M.~Branson, K.~Davis, M.~Kohler, J.~Meyer, M.~Henretty, R.~Morais,
  L.~Saunders, F.~Tyers, and G.~Weber.
\newblock Common voice: A massively-multilingual speech corpus.
\newblock In \emph{LREC}, 2020.

\bibitem[Asami et~al.(2017)Asami, Masumura, Yamaguchi, Masataki, and
  Aono]{asami2017domain}
T.~Asami, R.~Masumura, Y.~Yamaguchi, H.~Masataki, and Y.~Aono.
\newblock Domain adaptation of dnn acoustic models using knowledge
  distillation.
\newblock In \emph{ICASSP}, 2017.

\bibitem[Baevski and Auli(2019)]{baevski2018adaptive}
A.~Baevski and M.~Auli.
\newblock Adaptive input representations for neural language modeling.
\newblock In \emph{International Conference on Learning Representations}, 2019.

\bibitem[Balam et~al.(2020)Balam, Huang, Lavrukhin, Deng, Majumdar, and
  Ginsburg]{balam2020improving}
J.~Balam, J.~Huang, V.~Lavrukhin, S.~Deng, S.~Majumdar, and B.~Ginsburg.
\newblock Improving noise robustness of an end-to-end neural model for
  automatic speech recognition, 2020.

\bibitem[Barker et~al.(2018)Barker, Watanabe, Vincent, and
  Trmal]{barker2018fifth}
J.~Barker, S.~Watanabe, E.~Vincent, and J.~Trmal.
\newblock The fifth {CHiME} speech separation and recognition challenge:
  dataset, task and baselines.
\newblock In \emph{ICSA Speech}, 2018.

\bibitem[Boeddeker et~al.(2018)Boeddeker, Heitkaemper, Schmalenstroeer, Drude,
  Heymann, and Haeb-Umbach]{boeddeker2018front}
C.~Boeddeker, J.~Heitkaemper, J.~Schmalenstroeer, L.~Drude, J.~Heymann, and
  R.~Haeb-Umbach.
\newblock Front-end processing for the chime-5 dinner party scenario.
\newblock In \emph{CHiME5 Workshop, Hyderabad, India}, 2018.

\bibitem[Cieri et~al.(2004, 2005{\natexlab{a}})Cieri, Graff, Kimball, Miller,
  and Walker]{cieri2004fisher}
C.~Cieri, D.~Graff, O.~Kimball, D.~Miller, and K.~Walker.
\newblock Fisher english training speech parts 1 and 2 transcripts
  {LDC200\{4,5\}T19}.
\newblock \emph{Philadelphia: LDC}, 2004, 2005{\natexlab{a}}.

\bibitem[Cieri et~al.(2004, 2005{\natexlab{b}})Cieri, Miller, and
  Walker]{cieri2004fisher2}
C.~Cieri, D.~Miller, and K.~Walker.
\newblock Fisher english training speech parts 1 and 2 {LDC200\{4,5\}S13}.
\newblock \emph{Philadelphia: LDC}, 2004, 2005{\natexlab{b}}.

\bibitem[Collobert et~al.(2016)Collobert, Puhrsch, and
  Synnaeve]{collobert2016wav2letter}
R.~Collobert, C.~Puhrsch, and G.~Synnaeve.
\newblock Wav2letter: an end-to-end convnet-based speech recognition system.
\newblock \emph{arXiv preprint arXiv:1609.03193}, 2016.

\bibitem[Duchi et~al.(2011)Duchi, Hazan, and Singer]{duchi2011adaptive}
J.~Duchi, E.~Hazan, and Y.~Singer.
\newblock Adaptive subgradient methods for online learning and stochastic
  optimization.
\newblock \emph{Journal of machine learning research}, 12, 2011.

\bibitem[Fan et~al.(2020)Fan, Grave, and Joulin]{fan2019reducing}
A.~Fan, E.~Grave, and A.~Joulin.
\newblock Reducing transformer depth on demand with structured dropout.
\newblock In \emph{ICML}, 2020.

\bibitem[Fiscus et~al.(2007)]{rt03s}
J.~G. Fiscus et~al.
\newblock 2003 nist rich transcription evaluation data {LDC2007S10}.
\newblock \emph{Web Download. Philadelphia: LDC}, 2007.

\bibitem[Garofolo et~al.(1993)Garofolo, Graff, Paul, and
  Pallett]{garofolo1993csr}
J.~Garofolo, D.~Graff, D.~Paul, and D.~Pallett.
\newblock {CSR-I} ({WSJ0}) complete {LDC93S6A}.
\newblock \emph{Web Download. Philadelphia: LDC}, 1993.

\bibitem[Gemmeke et~al.(2017)Gemmeke, Ellis, Freedman, Jansen, Lawrence, Moore,
  Plakal, and Ritter]{gemmeke2017audio}
J.~F. Gemmeke, D.~P. Ellis, D.~Freedman, A.~Jansen, W.~Lawrence, R.~C. Moore,
  M.~Plakal, and M.~Ritter.
\newblock Audio set: An ontology and human-labeled dataset for audio events.
\newblock In \emph{2017 IEEE International Conference on Acoustics, Speech and
  Signal Processing (ICASSP)}, pages 776--780. IEEE, 2017.

\bibitem[Ghahremani et~al.(2017)Ghahremani, Manohar, Hadian, Povey, and
  Khudanpur]{ghahremani2017investigation}
P.~Ghahremani, V.~Manohar, H.~Hadian, D.~Povey, and S.~Khudanpur.
\newblock Investigation of transfer learning for {ASR} using {LF-MMI} trained
  neural networks.
\newblock In \emph{ASRU}, 2017.

\bibitem[Godfrey and Holliman(1993)]{godfrey1993switchboard}
J.~Godfrey and E.~Holliman.
\newblock Switchboard-1 release 2 {LDC97S62}.
\newblock \emph{Philadelphia: LDC}, 1993.

\bibitem[Graves et~al.(2006)Graves, Fern{\'a}ndez, Gomez, and
  Schmidhuber]{graves2006connectionist}
A.~Graves, S.~Fern{\'a}ndez, F.~Gomez, and J.~Schmidhuber.
\newblock Connectionist temporal classification: labelling unsegmented sequence
  data with recurrent neural networks.
\newblock In \emph{ICML}, 2006.

\bibitem[Gulati et~al.(2020)Gulati, Qin, Chiu, Parmar, Zhang, Yu, Han, Wang,
  Zhang, Wu, et~al.]{gulati2020conformer}
A.~Gulati, J.~Qin, C.-C. Chiu, N.~Parmar, Y.~Zhang, J.~Yu, W.~Han, S.~Wang,
  Z.~Zhang, Y.~Wu, et~al.
\newblock Conformer: Convolution-augmented transformer for speech recognition.
\newblock \emph{arXiv preprint arXiv:2005.08100}, 2020.

\bibitem[Hadian et~al.(2018)Hadian, Sameti, Povey, and
  Khudanpur]{hadian2018end}
H.~Hadian, H.~Sameti, D.~Povey, and S.~Khudanpur.
\newblock End-to-end speech recognition using lattice-free {MMI}.
\newblock In \emph{Interspeech}, 2018.

\bibitem[Han et~al.(2017)Han, Chandrashekaran, Kim, and Lane]{han2017capio}
K.~J. Han, A.~Chandrashekaran, J.~Kim, and I.~Lane.
\newblock The {CAPIO} 2017 conversational speech recognition system.
\newblock \emph{arXiv preprint arXiv:1801.00059}, 2017.

\bibitem[Heafield(2011)]{heafield2011kenlm}
K.~Heafield.
\newblock {KenLM}: Faster and smaller language model queries.
\newblock In \emph{Proceedings of the sixth workshop on statistical machine
  translation}. Association for Computational Linguistics, 2011.

\bibitem[Hernandez et~al.(2018)Hernandez, Nguyen, Ghannay, Tomashenko, and
  Est{\`e}ve]{hernandez2018ted}
F.~Hernandez, V.~Nguyen, S.~Ghannay, N.~Tomashenko, and Y.~Est{\`e}ve.
\newblock {TED-LIUM} 3: twice as much data and corpus repartition for
  experiments on speaker adaptation.
\newblock In \emph{SPECOM}, 2018.

\bibitem[Ko et~al.(2017)Ko, Peddinti, Povey, Seltzer, and
  Khudanpur]{ko2017study}
T.~Ko, V.~Peddinti, D.~Povey, M.~L. Seltzer, and S.~Khudanpur.
\newblock A study on data augmentation of reverberant speech for robust speech
  recognition.
\newblock In \emph{ICASSP}, 2017.

\bibitem[Kunze et~al.(2017)Kunze, Kirsch, Kurenkov, Krug, Johannsmeier, and
  Stober]{kunze2017transfer}
J.~Kunze, L.~Kirsch, I.~Kurenkov, A.~Krug, J.~Johannsmeier, and S.~Stober.
\newblock Transfer learning for speech recognition on a budget.
\newblock In \emph{ACL Workshop on Representation Learning for NLP}, 2017.

\bibitem[LDC and NIST(1994)]{linguistic1994csr}
LDC and M.~I.~G. NIST.
\newblock {CSR-II} ({WSJ1}) complete {LDC94S13A}.
\newblock \emph{Web Download. Philadelphia: LDC}, 1994.

\bibitem[LDC et~al.(2002)]{hub5}
LDC et~al.
\newblock 2000 hub5 english evaluation speech {LDC2002S09} and transcripts
  {LDC2002T43}.
\newblock \emph{Web Download. Philadelphia: LDC}, 2002.

\bibitem[Maciejewski et~al.(2020)Maciejewski, Wichern, McQuinn, and
  Le~Roux]{maciejewski2020whamr}
M.~Maciejewski, G.~Wichern, E.~McQuinn, and J.~Le~Roux.
\newblock {Whamr!}: Noisy and reverberant single-channel speech separation.
\newblock In \emph{ICASSP}, 2020.

\bibitem[Manohar et~al.(2017)Manohar, Povey, and Khudanpur]{manohar2017jhu}
V.~Manohar, D.~Povey, and S.~Khudanpur.
\newblock {JHU Kaldi} system for {Arabic} {MGB-3} {ASR} challenge using
  diarization, audio-transcript alignment and transfer learning.
\newblock In \emph{ASRU}, 2017.

\bibitem[Medennikov et~al.(2020)Medennikov, Korenevsky, Prisyach, Khokhlov,
  Korenevskaya, Sorokin, Timofeeva, Mitrofanov, Andrusenko, Podluzhny,
  et~al.]{medennikov2020stc}
I.~Medennikov, M.~Korenevsky, T.~Prisyach, Y.~Khokhlov, M.~Korenevskaya,
  I.~Sorokin, T.~Timofeeva, A.~Mitrofanov, A.~Andrusenko, I.~Podluzhny, et~al.
\newblock The stc system for the chime-6 challenge.
\newblock In \emph{CHiME 2020 Workshop on Speech Processing in Everyday
  Environments}, 2020.

\bibitem[Menne et~al.(2016)]{menne2016rwth}
T.~Menne et~al.
\newblock The {RWTH/UPB/FORTH} system combination for the 4th {CHiME} challenge
  evaluation.
\newblock In \emph{CHiME-4 Workshop}, 2016.

\bibitem[Panayotov et~al.(2015)Panayotov, Chen, Povey, and
  Khudanpur]{panayotov2015librispeech}
V.~Panayotov, G.~Chen, D.~Povey, and S.~Khudanpur.
\newblock Librispeech: an {ASR} corpus based on public domain audio books.
\newblock In \emph{ICASSP}, 2015.

\bibitem[Park et~al.(2019)Park, Chan, Zhang, Chiu, Zoph, Cubuk, and
  Le]{park2019specaug}
D.~S. Park, W.~Chan, Y.~Zhang, C.-C. Chiu, B.~Zoph, E.~D. Cubuk, and Q.~V. Le.
\newblock {SpecAugment}: A simple data augmentation method for automatic speech
  recognition.
\newblock In \emph{Interspeech}, 2019.

\bibitem[Povey et~al.(2011)Povey, Ghoshal, Boulianne, Burget, Glembek, Goel,
  Hannemann, Motlicek, Qian, Schwarz, et~al.]{daniel2011kaldi}
D.~Povey, A.~Ghoshal, G.~Boulianne, L.~Burget, O.~Glembek, N.~Goel,
  M.~Hannemann, P.~Motlicek, Y.~Qian, P.~Schwarz, et~al.
\newblock The {Kaldi} speech recognition toolkit.
\newblock In \emph{ASRU}, 2011.

\bibitem[Pratap et~al.(2019)Pratap, Hannun, Xu, Cai, Kahn, Synnaeve,
  Liptchinsky, and Collobert]{pratap2018wav2letter}
V.~Pratap, A.~Hannun, Q.~Xu, J.~Cai, J.~Kahn, G.~Synnaeve, V.~Liptchinsky, and
  R.~Collobert.
\newblock wav2letter++: The fastest open-source speech recognition system.
\newblock In \emph{ICASSP}, 2019.

\bibitem[Seltzer et~al.(2013)Seltzer, Yu, and Wang]{seltzer2013investigation}
M.~L. Seltzer, D.~Yu, and Y.~Wang.
\newblock An investigation of deep neural networks for noise robust speech
  recognition.
\newblock In \emph{ICASSP}, 2013.

\bibitem[{Shannon}(1949)]{shannon1949comm}
C.~E. {Shannon}.
\newblock Communication in the presence of noise.
\newblock \emph{Proceedings of the IRE}, 37\penalty0 (1):\penalty0 10--21,
  1949.

\bibitem[Sun et~al.(2018)Sun, Yeh, Ostendorf, Hwang, and Xie]{sun2018training}
S.~Sun, C.-F. Yeh, M.~Ostendorf, M.-Y. Hwang, and L.~Xie.
\newblock Training augmentation with adversarial examples for robust speech
  recognition.
\newblock In \emph{Interspeech}, 2018.

\bibitem[Synnaeve et~al.(2019)Synnaeve, Xu, Kahn, Likhomanenko, Grave, Pratap,
  Sriram, Liptchinsky, and Collobert]{synnaeve2019end}
G.~Synnaeve, Q.~Xu, J.~Kahn, T.~Likhomanenko, E.~Grave, V.~Pratap, A.~Sriram,
  V.~Liptchinsky, and R.~Collobert.
\newblock End-to-end {ASR}: from supervised to semi-supervised learning with
  modern architectures.
\newblock \emph{arXiv preprint arXiv:1911.08460}, 2019.

\bibitem[Sz{\"o}ke et~al.(2019)Sz{\"o}ke, Sk{\'a}cel, Mo{\v{s}}ner, Paliesek,
  and {\v{C}}ernock{\`y}]{szoke2019building}
I.~Sz{\"o}ke, M.~Sk{\'a}cel, L.~Mo{\v{s}}ner, J.~Paliesek, and
  J.~{\v{C}}ernock{\`y}.
\newblock Building and evaluation of a real room impulse response dataset.
\newblock \emph{IEEE Journal of Selected Topics in Signal Processing},
  13\penalty0 (4):\penalty0 863--876, 2019.

\bibitem[Szyma{\'n}ski et~al.(2020)Szyma{\'n}ski, Piotr, {\.Z}elasko, Morzy,
  Szymczak, {\.Z}y{\l}a-Hoppe, Banaszczak, Augustyniak, Mizgajski, and
  Carmiel]{szymanski2020we}
Szyma{\'n}ski, Piotr, P.~{\.Z}elasko, M.~Morzy, A.~Szymczak,
  M.~{\.Z}y{\l}a-Hoppe, J.~Banaszczak, L.~Augustyniak, J.~Mizgajski, and
  Y.~Carmiel.
\newblock Wer we are and wer we think we are.
\newblock \emph{arXiv preprint arXiv:2010.03432}, 2020.

\bibitem[Vaswani et~al.(2017)Vaswani, Shazeer, Parmar, Uszkoreit, Jones, Gomez,
  Kaiser, and Polosukhin]{vaswani2017attention}
A.~Vaswani, N.~Shazeer, N.~Parmar, J.~Uszkoreit, L.~Jones, A.~N. Gomez,
  {\L}.~Kaiser, and I.~Polosukhin.
\newblock Attention is all you need.
\newblock In \emph{Advances in Neural Information Processing Systems}, 2017.

\bibitem[Vinyals et~al.(2012)Vinyals, Ravuri, and Povey]{vinyals2012revisiting}
O.~Vinyals, S.~V. Ravuri, and D.~Povey.
\newblock Revisiting recurrent neural networks for robust {ASR}.
\newblock In \emph{ICASSP}, 2012.

\bibitem[Watanabe et~al.(2020)Watanabe, Mandel, Barker, Vincent, Arora, Chang,
  Khudanpur, Manohar, Povey, Raj, et~al.]{watanabe2020chime}
S.~Watanabe, M.~Mandel, J.~Barker, E.~Vincent, A.~Arora, X.~Chang,
  S.~Khudanpur, V.~Manohar, D.~Povey, D.~Raj, et~al.
\newblock Chime-6 challenge: Tackling multispeaker speech recognition for
  unsegmented recordings.
\newblock \emph{arXiv preprint arXiv:2004.09249}, 2020.

\bibitem[Wenzek et~al.(2020)Wenzek, Lachaux, Conneau, Chaudhary, Guzm{\'a}n,
  Joulin, and Grave]{wenzek2020ccnet}
G.~Wenzek, M.-A. Lachaux, A.~Conneau, V.~Chaudhary, F.~Guzm{\'a}n, A.~Joulin,
  and {\'E}.~Grave.
\newblock Ccnet: Extracting high quality monolingual datasets from web crawl
  data.
\newblock In \emph{Proceedings of The 12th Language Resources and Evaluation
  Conference}, pages 4003--4012, 2020.

\bibitem[Woodland et~al.(1994)Woodland, Odell, Valtchev, and
  Young]{woodland1994large}
P.~C. Woodland, J.~J. Odell, V.~Valtchev, and S.~J. Young.
\newblock Large vocabulary continuous speech recognition using {HTK}.
\newblock In \emph{ICASSP}, 1994.

\bibitem[Zhou et~al.(2020)Zhou, Michel, Irie, Kitza, Schl{\"u}ter, and
  Ney]{zhou2020rwth}
W.~Zhou, W.~Michel, K.~Irie, M.~Kitza, R.~Schl{\"u}ter, and H.~Ney.
\newblock The {RWTH ASR} system for {TED-LIUM} release 2: Improving hybrid
  {HMM} with specaugment.
\newblock In \emph{ICASSP}, 2020.

\end{thebibliography}

\end{document}